\documentclass[letterpaper, 10 pt, conference]{ieeeconf}
\IEEEoverridecommandlockouts                              

\usepackage[T1]{fontenc}
\usepackage{times}
\usepackage{cite}
\usepackage{amsmath,amssymb,amsfonts}
\usepackage{graphicx}
\usepackage{textcomp}
\usepackage{xcolor}
\usepackage{booktabs}
\usepackage{multirow}
\usepackage{url}

\makeatletter
\let\NAT@parse\undefined
\makeatother
\usepackage[colorlinks=true,linkcolor=blue,citecolor=red,urlcolor=cyan,
            pdftitle={HEROIC}]{hyperref}

\graphicspath{{figures/}}
\makeatletter
\def\bstctlcite{\@ifnextchar[{\@bstctlcite}{\@bstctlcite[@auxout]}}
\def\@bstctlcite[#1]#2{\@bsphack
  \@for\@citeb:=#2\do{%
    \edef\@citeb{\expandafter\@firstofone\@citeb}%
    \if@filesw\immediate\write\csname #1\endcsname{\string\citation{\@citeb}}\fi}%
  \@esphack}
\makeatother
\newcommand{\rmax}{R_{\max}}
\newcommand{\msg}[1]{\texttt{[#1]}}
\newcommand{\ind}[1]{\mathbf{1}\!\left[#1\right]}
\begin{document}
\bstctlcite{IEEEexample:BSTcontrol}  

\title{\LARGE \bf
HEROIC: Heterogeneous Evidential Reasoning for Open-Vocabulary Identification and Cross-Robot Collaboration
}

\author{Mihir Chauhan$^{1}$, Aarav Jain$^{1}$, Addison Zucek$^{1}$, Manmeet Dang$^{1}$, Damon Conover$^{2}$, Aniket Bera$^{1}$%
\thanks{\raggedright $^{1}$Mihir Chauhan, Aarav Jain, Addison Zucek, Manmeet Dang, and Aniket Bera are with the IDEAS Lab, Department of Computer Science, Purdue University, West Lafayette, IN, USA. %
        {\tt\small \{\href{mailto:chauhanm@purdue.edu}{chauhanm}\allowbreak,\href{mailto:aniketbera@purdue.edu}{aniketbera}\allowbreak\}@\allowbreak purdue\allowbreak.edu}}
\thanks{\raggedright $^{2}$Damon Conover is with the DEVCOM Army Research Laboratory, Adelphi, MD, USA. %
        {\tt\small \{\href{mailto:chauhanm@purdue.edu}{damon.m.conover.civ@army.mil}\}}}%
}

\maketitle

\begin{abstract}
Multi-agent heterogeneous air-ground robot teams are attractive for open world search, with applications for reconnaissance, urban search and rescue missions (USAR), disaster response and recovery, and hazardous environments. These two platforms have different failure modes: aerial robots cover ground quickly but cannot resolve small or occluded targets from altitude, while ground robots can identify objects-of-interest, such as people or hazardous objects, at close range but cover less area. Existing language-tasked teams either have roles fixed prior, or have a language model assign them from hand-written capability tags, so the team is unable to know when within a mission an asset is no longer useful. We present HEROIC, a decentralized heterogeneous multi-agent open-vocabulary search coordination framework that requires agents to communicate in natural language only. HEROIC's initial agent role assignment is derived from sensor properties and a scale law to determine whether targets can be detected with a high confidence. From the mission's natural language prompt alone, this law assigns aerial flight altitudes and sweep spacing. When this calculated height falls below the altitude for safe flight, aerial agents re-task themselves from searcher to aerial triage, escort, and route guide for ground agents. Both robots maintain an evidential belief over the search area (bearing rays for positive evidence, a log-odds posterior for negative evidence) and gate any arrival on close-range verification. In full-stack experiments, HEROIC reaches the target 84\% of the time across all 6 scenes, compares to 35-54\% for vision-language frontier baselines, frontier-based search, lawnmower, and random-walk running the same perception, all while being 2-4x sooner to arrive at the target.
\end{abstract}

\section{Introduction}
People trapped in a destroyed warehouse, a hiker lost under a dense forest canopy, and an injured worker in a smoke-filled factory all have one thing in common: the fastest way to reach them is with an aerial robot, but usually a UAV cannot safely reach them. Ground robots can carry supplies, lift rubble and see at close range with less occlusion. Wide-area coverage and close-range identification are properties of two different platforms, and the value of air-ground teams lies in moving work from one platform to the other effectively~\cite{delmerico2019rescue,queralta2020collaborative,murphy2014disaster}. Recent work with language-tasked teams~\cite{cladera2025airground,miller2022stronger,chen2024scalable,arul2024sayconav,ravichandran2025spine} uses open-vocabulary perception with large language model (LLM) planners for reasoning on a natural-language mission.

\begin{figure}[htbp]
    \centerline{\includegraphics[width=3.5in]{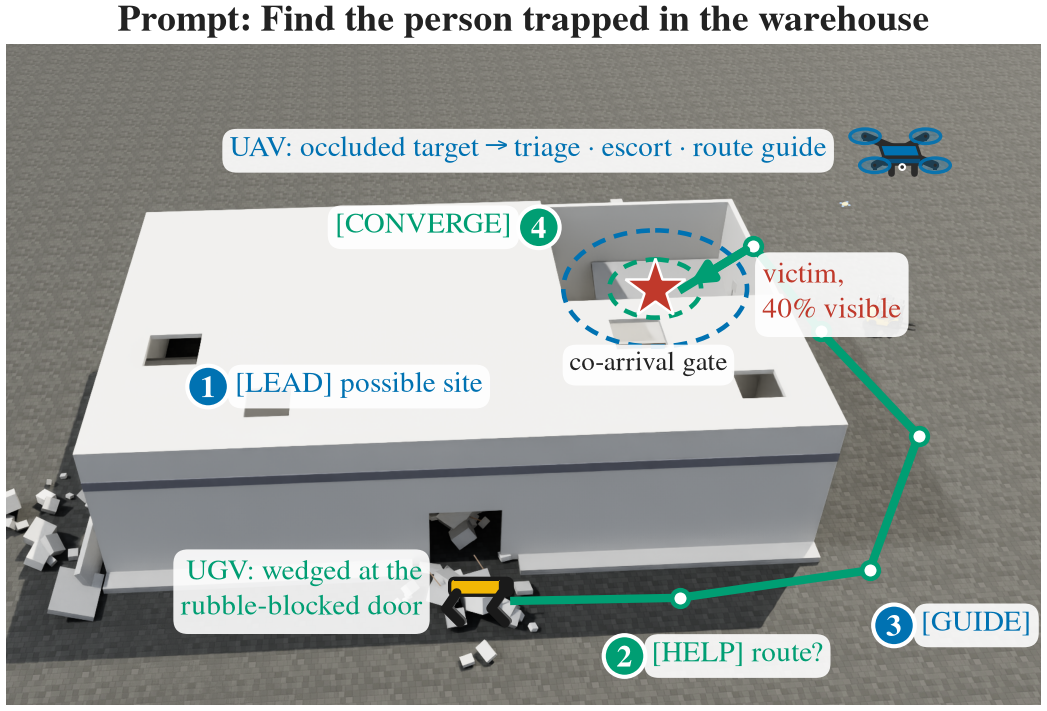}}
    \caption{HEROIC deployed on a mission: a person trapped in a collapsed warehouse. The aerial robot cannot resolve a 40\%-visible person from a safe altitude, so it re-tasks itself from searcher to helper: it triages rubble-blocked bays and finds a point of entry for the ground robot. Overlaid: the robots' exchange.}
    \label{fig:cover}
\end{figure}

However, they do not yet provide a framework where roles are variable, such as in Fig.~\ref{fig:cover}, where the warehouse roof occludes the insides and the aerial robot must switch from searching to helping the ground robot navigate. Existing solutions fix roles in advance, such as Cladera et al.~\cite{cladera2025airground} where the ground vehicle is the mission executor while the aerial vehicle provides semantic graphs. However, sensor limits can drastically change the effectiveness of a role: an aerial camera cannot resolve humans underneath rubble.

An open-vocabulary detector requires a target to occupy a minimum pixel area to be separated from noise. We measured this threshold for our detector; the resulting range bound $R_{max}=f_yh_{vis}/px_{min}$ is the primitive from which the team's behavior is derived (Sec.~\ref{sec:law}). Given only a natural-language mission prompt, $R_{max}$ yields the initial cruise altitude for the UAVs, sweep pacing, and a blind bit that bit re-tasks the aerial robot from searcher to \textit{aerial triage}, \textit{escort}, and \textit{route guide} (Sec.~\ref{sec:roles}). Both robots maintain an evidential belief that distinguishes positive evidence from negative evidence, and no agent declares an arrival without close-range verification by its own detector and a vision-language model (VLM). Coordination is strictly in natural language over a peer-to-peer channel, which keeps the team's strategy human-readable and lets an operator redirect it with a single message.

Our contributions are fourfold:
\begin{enumerate}
    \item \textbf{Scale-law role assignment:} a formulation that turns one measured sensor property, the pixel extent below which the detector cannot separate target from clutter, into the UAV's altitude, sweep pacing, and a per-mission, per-target role assignment for an air-ground team.
    \item \textbf{Evidential belief and arrival gate:} a belief that keeps positive evidence as triangulated bearing rays and negative evidence as a log-odds posterior, together with a verification gate, committed hold and veto that let the team commit to or retract target locations without ground truth.
    \item \textbf{Two-rate decentralized architecture:} a slower LLM loop handles strategy and all natural-language coordination, while a 1 Hz reflex layer handles investigation, waypoint sweeps, and the detector.
    \item \textbf{Paired evaluation across six scene tiers:} every baseline runs inside the same stack with one function swapped; full-stack Isaac~Sim missions on procedurally generated scenes are the primary comparison, and a detector-calibrated abstraction provides $n=300$ paired component ablations.
\end{enumerate}

On the tiers where the scale law determines that aerial robots can see the target, HEROIC is comparable to the strongest baseline; on tiers where it cannot, where collaboration is more beneficial, HEROIC is the only method that does not collapse.

\section{Related Work}
\label{sec:related}
\subsection{Heterogeneous UAV-UGV Teams }
Kulkarni et al. \cite{9812401} deploy a UGV and UAV team for autonomous subterranean exploration, assigning roles over a map. De Petrillo et al. \cite{9448603} plan where a UAV should fly to balance coverage against localizability relative to a UGV anchor. Both coordinate by sharing map data; HEROIC's robots coordinate in natural language and divide labor by what each can physically resolve. Cladera et al.~\cite{cladera2025airground} present the most closely
related system: a UAV-UGV team that accomplishes language-specified 
missions in unknown environments by building semantic-metric maps 
online and sharing them between robots. Miller et al.~\cite{miller2022stronger} 
similarly use the UAV to provide semantic context that the UGV uses 
for planning. In both systems the aerial role is fixed before the 
mission begins.

\subsection{Large Language Models for Multi-Robot Coordination}
RoCo \cite{10610855} handles multi-robot manipulation as a conversation among LLM agents that negotiate toward a joint plan, which a centralized motion planner then validates. Chen et al. \cite{chen2024scalable} study whether LLM planners should be centralized or decentralized, while SMART-LLM \cite{10802322} breaks down natural language instruction into smaller tasks, creates teams, and allocates work. These methods assume a known environment and object set and treat the LLM as a one-shot problem solver; our per-agent LLM policies run in an unknown scene and relay what they learn to each other in natural language.

\subsection{Open-Vocabulary Object Detection and Vision-Language Navigation}

VLFM \cite{yokoyama2024vlfm} uses a vision-language model (VLM) to handle object search, and Co-NavGPT \cite{11302789} uses a VLM as a global planner to assign team goals over a global map. HEROIC also pairs open-vocabulary detection with depth, but couples it to language-coordinated heterogeneous agents. 
\subsection{Semantic Mapping and Shared Memory}
ConceptGraphs \cite{10610243} builds an open-vocabulary 3D scene graph from RGB images, with objects as nodes and relations as edges, and plans over it. HD-CCSOM \cite{9981756} fuses each robot's semantic occupancy map into a shared one to reconcile labels across viewpoints. Both require continuous exchange of dense map data; in HEROIC each agent keeps a lightweight object-centric memory and shares detections, positions and hypotheses only as natural-language messages.

\section{Problem Formulation}
\label{sec:problem}
We formulate the task of language-specified search by an air-ground team as a Decentralized Partially Observable Markov Decision Process with Communication (Dec-POMDP-Com)~\cite{bernstein2002complexity,oliehoek2016concise}. A team of one UAV with a gimballed RGB-D camera and one legged UGV with a body-mounted RGB-D camera receives a natural language goal $\mathcal{G}$ and an operator-supplied search disk $\mathcal{D} \subset \mathbb{R}^2$ of radius $R_s$. The scene is entirely unknown: the agents never see the object list or ground truth. All agents run identical perception and agent loops, communicating only in natural language over a peer-to-peer channel; there is no central planner and no shared state.

Formally, the Dec-POMDP is the tuple
$\mathcal{M} = \big(\mathcal{I}, \mathcal{S}, \{\mathcal{A}^i\}, T, R, \{\Omega^i\}, O, \Sigma, H\big)$,
where $\mathcal{I} = \{\mathrm{uav}, \mathrm{ugv}\}$ is the set of agents, $\mathcal{S}$ is the state space, $\mathcal{A}^i$ the action space of agent $i$ with joint action $\mathbf{a} = (a^{\mathrm{uav}}, a^{\mathrm{ugv}})$, $T(s' \mid s, \mathbf{a})$ the transition function, $R(s, \mathbf{a})$ the shared reward, $\Omega^i$ the observation space of agent $i$, $O(\mathbf{o} \mid s', \mathbf{a})$ the joint observation function, $\Sigma$ the message alphabet, and $H = T_{\max}/\Delta t$ the finite horizon with $T_{\max} = 900$~s of simulation time. Following the Dec-POMDP-Com construction, a message is part of the sending agent's action and part of the receiving agent's next observation, so communication is an action the policy makes. $T$ is the simulator (rigid-body dynamics and collisions, a stall when the UGV wedges against unmapped geometry) plus message delivery; $\mathcal{E}$, $\mathbf{p}^\star$ and $\nu$ are static within an episode and drawn from a seeded generator across episodes.

\subsection{State, Actions and Communication}
The state $s_t \in \mathcal{S}$ at time $t$ comprises the robot poses, $\mathbf{x}^{\mathrm{uav}}_t \in \mathbb{R}^3 \times SO(3)$ including gimbal angles and $\mathbf{x}^{\mathrm{ugv}}_t \in SE(2)$; the scene $\mathcal{E}$: terrain, buildings, rubble, canopy and a set of objects $\{(\ell_k, \mathbf{p}_k, h_k)\}$ with open-vocabulary label, position and height, of which one or more carry the goal class; the targets' position(s) $\mathbf{p}^\star \in \mathbb{R}^2$ and \emph{visible fraction} $\nu \in (0,1]$, the share of height not hidden by rubble, walls or canopy; and the messages in flight on the peer channel plus a dwell timer $\delta_t \ge 0$ recording how long both robots have been simultaneously inside their arrival radii of the victim. Nothing in $\mathcal{E}$, $\mathbf{p}^\star$, $\nu$ or $\delta_t$ is available to the agents. The scale of the goal object, $h$, is modeled on the goal noun in $\mathcal{G}$, and $\nu$ is a property of the mission brief for occluded targets.

An action of agent $i$ is $a^i_t = (\mathbf{g}^i_t, \gamma^i_t, m^i_t) \in \mathcal{A}^i$: a motion target $\mathbf{g}^i_t$ (a waypoint or a route), a camera command $\gamma^i_t$ (gimbal or heading sweep) and a message $m^i_t \in \Sigma \cup \{\varnothing\}$ broadcast on the peer-to-peer channel. The alphabet $\Sigma$ is unrestricted natural language in which every message opens with one of a small set of tags (Sec.~\ref{sec:protocol}). Each agent acts on a decentralized policy over its own action--observation history,
\begin{equation}
    a^i_t \sim \pi^i\big(\cdot \mid h^i_t\big), \qquad h^i_t = (o^i_1, a^i_1, \ldots, a^i_{t-1}, o^i_t),
\end{equation}
with joint policy $\boldsymbol{\pi} = (\pi^{\mathrm{uav}}, \pi^{\mathrm{ugv}})$; everything one agent knows about the other arrives through $\Sigma$.

\subsection{Observation Model}
Agent $i$ cannot observe the state space $s_t$, rather at each frame it receives
\begin{equation}
    o^i_t = \big(\hat{\mathbf{x}}^i_t,\; \mathcal{Y}^i_t,\; \mathbf{d}^i_t,\; \mathcal{C}^i_t\big) \in \Omega^i,
\end{equation}
its own pose estimate, a set of open-vocabulary detections $\mathcal{Y}^i_t = \{(\ell_k, b_k, c_k)\}$ with label, bounding box and confidence, a depth image $\mathbf{d}^i_t$, and the messages $\mathcal{C}^i_t \subset \Sigma$ delivered since the last frame. The detector is the source of both partial observability and asymmetry between the platforms. Its output for an object of visible height $h_{vis} = \nu h$ at slant range $R$ from a camera with focal length $f_y$ depends on the pixel extent the object subtends,
\begin{equation}
    px(R) = \frac{f_y\, h_{vis}}{R},
    \label{eq:px}
\end{equation}
and we model the probability of a goal-class detection, given that the object is in the field of view, as a monotone function $\eta(px)$ that is at chance below a threshold $px_{\min}$ and reliable above it (measured in Sec.~\ref{sec:law}). Because $px$ falls with range, Eq.~\eqref{eq:px} induces a maximum resolvable range $\rmax$ for every goal class, and from an altitude $a$ the ground disk in which the target can be detected has radius $\sqrt{\rmax^2 - a^2}$, which may be empty; this is what makes the aerial platform useful for some targets and not others.

Detections are also unreliable in a second way: a permissive open-vocabulary detector produces tens of goal-class phantoms per mission, and depth back-projection of a far detection has a median planar error of 10--18~m beyond a range of 10~m, compared to 1--2~m for the closest approach of the bearing ray. Evidence thus has a trustworthy \emph{bearing} but untrustworthy \emph{range} and \emph{label}, and the belief of Sec.~\ref{sec:belief} is built around this distinction.

\subsection{Reward and Objective}
If we let $r_{\mathrm{uav}} = 10$m and $r_{\mathrm{ugv}} = 5$m be the arrival radii and $\tau_d = 10$~s the dwell. The mission is complete at the first time both robots are simultaneously within their radii of the victim and remain there for the dwell,
\begin{equation}
    T^\star = \min\Big\{ t : \big\|\mathbf{x}^i_{t'} - \mathbf{p}^\star\big\| \le r_i \;\; \forall i \in \mathcal{I},\; \forall t' \in [t - \tau_d, t] \Big\}
    \label{eq:success}
\end{equation}
and the mission is a success if $T^\star \le T_{\max}$. The shared reward charges one time step for every step until completion,
$R(s_t, \mathbf{a}_t) = -\Delta t \cdot \ind{\, \delta_t < \tau_d \,}$, so the undiscounted return is $J(\boldsymbol{\pi}) = -\mathbb{E}\big[\min(T^\star, T_{\max})\big]$. A failed episode costs the full budget, so maximizing $J$ maximizes success probability and, when successful, minimizes time-to-find. Co-arrival is stricter than the UGV-only criterion of Cladera et al.~\cite{cladera2025airground} and appropriate for triage: the air asset must be on station as relay and overwatch when the ground robot reaches the target. On the multi-victim warehouse tier every victim is scored under the same gate and one trapped victim is designated primary.

Solving a Dec-POMDP optimally is NEXP-complete~\cite{bernstein2002complexity}, and here $\mathcal{S}$ is a procedurally generated 3D scene and $\Sigma$ is natural language. HEROIC is a structured decentralized policy: it exploits the structure of $O$ in \eqref{eq:px} to decide each agent's role, maintains a per-agent belief, and uses $\Sigma$ for information the partner's belief cannot supply. The reward is an evaluation objective only, not a learning signal.

We report \textbf{success rate}, the fraction of episodes with $T^\star \le T_{\max}$ under \eqref{eq:success}; \textbf{time-to-find}, $T^\star$ censored at $T_{\max}$ for failures, whose mean is $-J$; and, for every contrast, the \textbf{discordant pairs}: the number of seeds one method solves and the other does not, in both directions. 

\begin{figure*}[t]
  \centering
  \includegraphics[width=\textwidth]{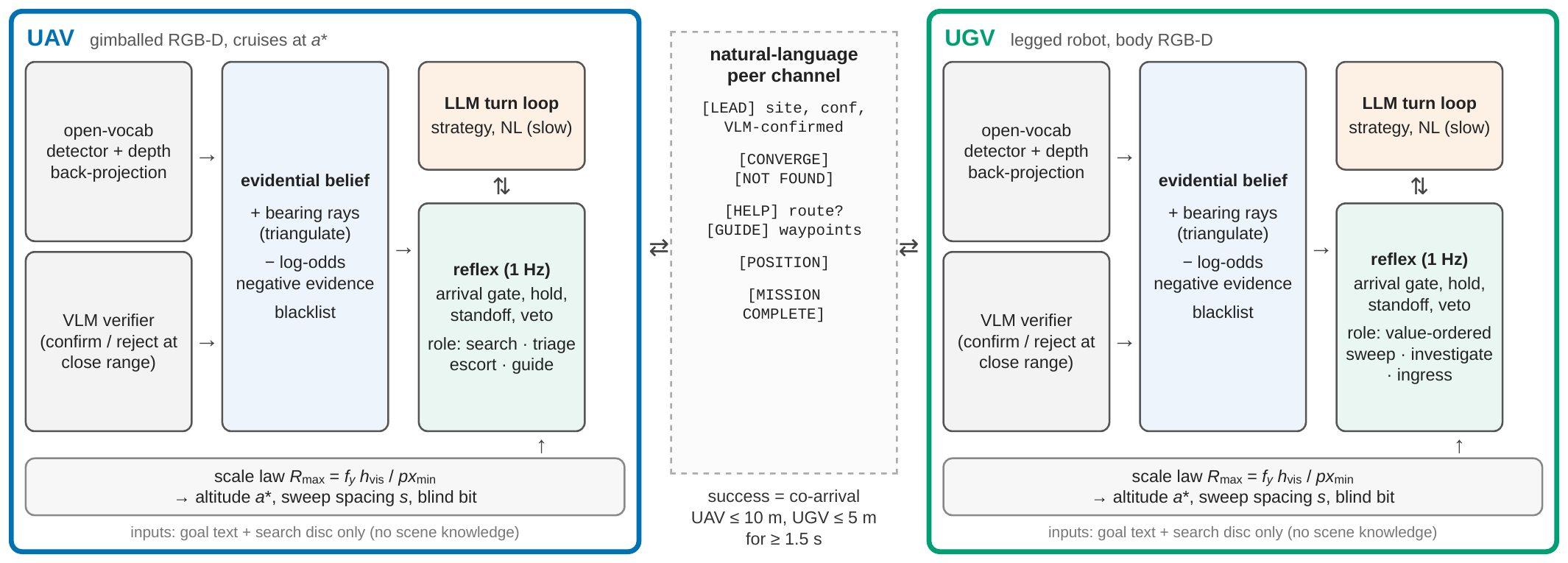}
  \caption{HEROIC Architecture: Each robot runs the same stack: open-vocabulary detection with depth back-projection and VLM verification feed an evidential belief; a slower LLM turn loop handles strategy and language while a 1~Hz reflex owns the arrival gate, committed hold, standoff and veto. The scale law, evaluated once per mission from the goal text, sets the UAV's altitude, sweep spacing and the blind bit that selects its role. }
  \label{fig:system}
\end{figure*}

\section{Methodology}
\label{sec:method}
Each robot runs the same decentralized stack (Fig.~\ref{fig:system}).

\subsection{The Scale Law}
\label{sec:law}

\begin{figure}[t]
  \centering
  \includegraphics[width=\linewidth]{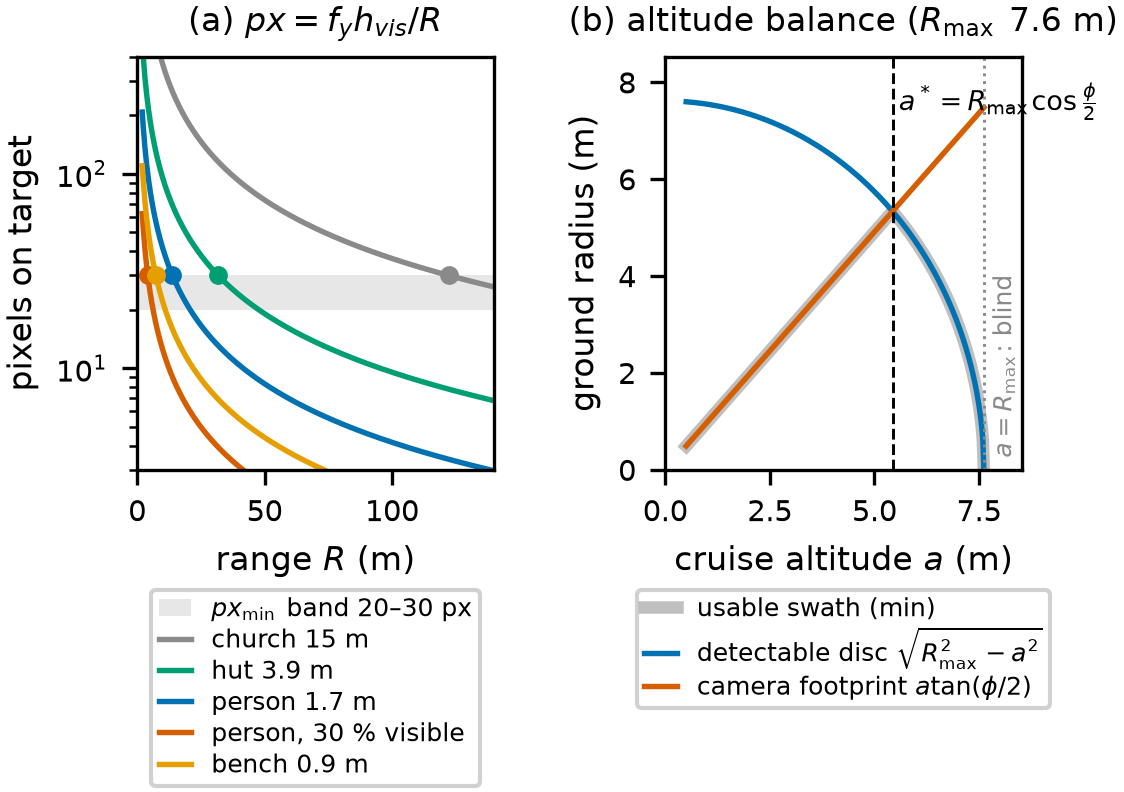}
  \caption{(a) Pixels on target versus range for the goal classes used in this paper; dots mark $\rmax$ at $px_{\min}=30$. (b) Detectable ground disk $\sqrt{\rmax^2-a^2}$ versus altitude $a$, zero at $a=\rmax$; balancing it against the camera footprint gives the cruise altitude $a^*$ of~\eqref{eq:alt}.}
  \label{fig:law}
\end{figure}

An open-vocabulary detector requires the object of interest to cover enough pixels to be distinguishable from clutter. Inverting \eqref{eq:px} at the threshold $px_{\min}$ gives the maximum range at which a target of expected visible height $h_{vis}$ can be resolved:
\begin{equation}
  \rmax = \frac{f_y\, h_{vis}}{px_{\min}}, \qquad h_{vis} = \nu\, h
  \label{eq:law}
\end{equation}
We measured $px_{\min}$ on the deployed detector using 64k detections matched to ground truth: a frozen-CLIP false-positive discriminator~\cite{radford2021clip} is at chance ($\approx 0.51$) below 20px and reliable by 30px, so $px_{\min} = 30$. With the runtime intrinsics ($f_y = 244.3$~px at $640\times480$) this gives $\rmax = 122$~m for a 15~m church, 13.8~m for a standing person, 7.6~m for a person 55\% visible in rubble and 4.2~m at 30\% visible (Fig.~\ref{fig:law}a). The height $h$ is looked up from the mission prompt noun in a 40-entry table (otherwise estimated by the LLM) and the visible fraction $\nu$ is estimated from the mission brief, and lowered if nothing is found at 50\% visible after a sweep of the search area; nothing in~\eqref{eq:law} depends on the scene instance.

\subsection{Scale-Derived Search Parameters}
\label{sec:derived}

Given $\rmax$, the following quantities are computed once per mission and shared by every method we evaluate.

\textbf{Cruise altitude.} Since detection depends on slant range, from altitude $a$ the ground disk in which the target is detectable has radius
$d(a) = \sqrt{\rmax^2 - a^2}$. The obvious choice $a = \rmax$ is degenerate ($d = 0$). We balance the detectable disk against the camera's ground footprint $a \tan(\phi/2)$, where $\phi$ is the vertical field of view; the two are equal at
\begin{equation}
  a^* = \rmax \cos(\phi/2)
  \label{eq:alt}
\end{equation}
which maximizes the smaller of the two and hence the usable swath seen in Fig.~\ref{fig:law}b. The commanded altitude is $a^*$ clamped to a flight-safety band, $a = \min(a_{\max}, \max(a_{\min}, a^*))$ with $[a_{\min}, a_{\max}] = [6, 25]$~m, and on canopy tiers lifted to the measured canopy height plus a 4~m clearance whenever $a^*$ allows it, with the camera at a $60^\circ$ decline so that it images the canopy gaps beneath the vehicle.

\textbf{Sweep spacing.} Consecutive swaths must stay inside the detectable range with overlap, so
$s = \max(4~\text{m},\, 0.75\, \rmax)$: small targets imply tight sweeps. The same $s$ spaces the ground robot's sweeps.

\textbf{Ground sensing radius.} The UGV marks cells within $\min(\rmax, 15~\text{m})$ as observed, so that ``explored'' means ``explored at a range at which the target was resolvable''.

\textbf{The blind bit.} The aerial robot is \emph{blind} for the current target if, at the lowest altitude it can fly, the detectable disk is smaller than one sweep,
\begin{equation}
  \beta = \ind{\, d(a) < 2~\text{m} \,}
  \label{eq:blind}
\end{equation}
When $\beta = 1$, every aerial sweep is noisy false positives, so this bit re-tasks the team (Sec.~\ref{sec:roles}).

\subsection{Evidential Belief}
\label{sec:belief}

Because bearing is reliable and range and label are not (Sec.~\ref{sec:problem}), the belief separates two kinds of evidence over the search area, discretized into a grid of cells $c$ with world position $\mathbf{w}_c$.

\textbf{Positive evidence as bearing rays.} Every mission-critical detection is fed as a ray $(\mathbf{o}, \hat{\mathbf{d}})$ from the camera origin through the center of the bounding-box to a ray tracker that associates rays across frames and poses. A track becomes \emph{committable} only when it is triangulated from at least $n_{\min} = 2$ distinct observer poses at least $b_{\min} = 5$~m apart with a triangulation uncertainty $\sigma \le 4$~m. A goal ray also adds a small, capped log-odds ridge along its ground projection so that an un-triangulated far sighting still pulls the robot's bearing,
\begin{equation}
  L(c) \leftarrow L(c) + \kappa(\mathrm{conf}) \exp\!\Big(-\frac{\rho_\perp(c)^2}{2\sigma_\perp^2}\Big),\quad
  \rho_\parallel(c) \in [3, 45]~\text{m},
  \label{eq:ridge}
\end{equation}
where $\rho_\parallel$ and $\rho_\perp$ are the along-ray and perpendicular distances of the cell, $\sigma_\perp = 3$~m and $\kappa = 0.30\,\mathrm{clip}(\mathrm{conf}/0.4, 0.2, 1)$. This guarantees that no single ray can be committable.

\textbf{Negative evidence as a log-odds posterior.} A cell observed at close range with no goal detection is a Bayesian miss. With $p_{det}(r)$ the measured probability of detecting the target at range $r$ when it is present, the update for every cell in the observed mask is
\begin{align}
  L(c) &\leftarrow L(c) + \log\big(1 - p_{det}(r_c)\big), \label{eq:neg}\\
  p_{det}(r) &= \mathrm{clip}\Big(p_0\, \mathrm{clip}\big(1.5 - \tfrac{r}{30}, 0.15, 1\big), 0.02, 0.95\Big), \nonumber
\end{align}
with $p_0 = 0.6$ from the detection model, so that farther cells are less reliably ``empty''. The normalized posterior $P(c) \propto \sigma(L(c))$ over the disk then decays phantom neighborhoods once a robot stands in them, and probability mass is in unobserved cells. Non-goal detections (rubble, vehicles, barns, trails) feed a separate context tracker that doesn't affect goal posterior.

\textbf{Blacklist.} Sites that the arrival gate has rejected enter a blacklist with a 6~m footprint that both robots honour; a site a robot could not get within 4~m of is reported \msg{UNREACHED} and is \emph{not} blacklisted, as an unreached site is not inspected.

\subsection{Arrival Gate, Committed Hold and Veto}
\label{sec:gate}

The \textbf{arrival gate} declares that it is ``on target'' only when (i) the robot is within the arrival radius of mission-critical cluster, (ii) the cluster is supported by its own close-range detections from at least two poses, (iii) the VLM verifier, run on the robot's own frame, has not rejected it, and, for mission completion, (iv) the partner is also within radius. Close-range detections are themselves gated by \eqref{eq:px}: a detection whose box height is below $px_{\min}$ is not admitted as evidence for commitment.

A verified arrival engages a \textbf{committed hold} announced with a \msg{CONVERGE} message. The hold parks the robot at a 2.5~m standoff on the ray from the target so that its own proximity cannot trigger a veto. The hold re-estimates its target from fresh evidence every tick and is vetoed by one of three: partner co-location without success (20~s), a mission-clock cap, or a \textbf{nothing-detected veto}, which fires after 30~s on station with zero fresh goal-class evidence within 6~m, after turning to face the target twice. The veto releases the hold, blacklists the location and broadcasts \msg{NOT FOUND}.

\subsection{Role Assignment Derived from the Law}
\label{sec:roles}

The aerial robot's role $\rho^{\mathrm{uav}} \in \{\textsc{searcher}, \textsc{blind}\}$ is a function of the blind bit, $\rho^{\mathrm{uav}} = \textsc{blind}$ $\Leftrightarrow \beta = 1$, evaluated per mission from~\eqref{eq:law}--\eqref{eq:blind} and never asserted in a prompt.

\textbf{Searcher} ($\beta = 0$). Cruise at $a$, sweep at spacing $s$, broadcast triangulated goal-class leads. Leads that the UAV's own VLM has confirmed are tagged \texttt{VLM-confirmed} and ranked first by the ground robot, without the credibility discount that a label otherwise accrues after rejections.

\textbf{Blind} ($\beta = 1$). Three complementary behaviours, each addressing a different way the blind regime fails.

\emph{Aerial triage.} The UAV enumerates \emph{container} classes for the goal (rubble, damaged building, vehicle, barn, hut, etc) from altitude, where they are resolvable by \eqref{eq:law} even when the victim is not, and flags one site at a time on a 6~m grid at a 25~s cadence as \msg{LEAD}. Container weights are not hand-set: an LLM reads the mission brief once and returns a cue prior over the detector vocabulary (hiker: trail 1.7, backpack 1.6; farm: smoke 1.8, barn 1.8), so the same code ranks rubble on the USAR tier and barns on the farm tier.

\emph{Escort.} The UAV shadows the ground searcher at an 8~m standoff, inside its own 10~m arrival radius, so the co-arrival gate \eqref{eq:success} can fire the moment the UGV verifies the victim.

\emph{Route guide.} The victim on the rubble and warehouse tiers is \emph{inside} a structure, and the last ten metres are a geometry problem rather than a detection problem. When the ground robot asks \msg{HELP} (issued when its 75~s refine window expires more than 4~m from a lead or when its route stalls twice), the UAV flies over the site, hovers while its depth camera builds a 0.5~m height map (per-cell maximum height, 0.4~m obstacle threshold), plans a shortest path from the UGV's pose to the reachable cell closest to a 3.5~m standoff from the target with costs for unknown, wall-adjacent and canopy cells, and answers \msg{GUIDE} with the waypoints. The ground robot treats the waypoints as advice: it joins them with its own planner and stall watchdog, and re-asks at most once per site.

\subsection{Ground Search Policy}
\label{sec:ground}

The ground robot's idle behaviour is a \emph{value-ordered sweep}: a Victor Sierra pattern over the disk at spacing $s$ whose waypoints are visited in order of an observation-conditioned salience prior with a distance penalty and a 40~m travel horizon. The salience prior is a heatmap over the disk conditioned on the goal-text embedding and the robot's own detections, so that cells near cue objects rank first. The sweep is pre-empted at every tick by two events: \emph{lead investigation}, which drives to the best unresolved goal-class cluster and rules it out if the gate does not verify within 75~s of arriving, and a partner \msg{CONVERGE}, which drives to the announced site. The ground robot also builds its own depth height map to route around walls no detector labels, marks the cell ahead of it blocked whenever a route stalls, and ends goal-directed routes at the reachable cell within 4.5~m of the target so that a victim behind a wall can be reached from outside it.

\subsection{Control Loop Architecture}
\label{sec:arch}

Each agent runs two loops on its own simulator socket.

\textbf{LLM turn loop.} Seconds to minutes per turn on a locally served model. It sees mission state, task allocation and peer messages each turn and acts through a narrow tool set: direct move, next frontier, coverage waypoint, gimbal sweep, scan area, report finding, broadcast, strategize. It handles strategy and all natural-language coordination, and is locked out whenever the reflex holds a motion target, so two planners never fight over one controller. After 80 turns the agent degrades to reflex-only.

\textbf{Reflex loop.} 1~Hz, no LLM. It owns peer-message intake, route driving with a stall watchdog and wedge back-off, close-range refinement with a micro-orbit for pose diversity, the committed hold and its release, the sweep, and the triage, escort, guide and rendezvous behaviours of Sec.~\ref{sec:roles}.

The split exists because the endgame is cadence-limited: with convergence handled in the turn loop the effective UGV endgame speed was 0.05--0.2~m/s and half the failures of an early run were near-misses with the UAV parked within 5~m of the target. 

\subsection{Language Protocol}
\label{sec:protocol}

All inter-robot traffic is plain language with a leading tag (Fig.~\ref{fig:system}): \msg{POSITION} heartbeats every 8~s, \msg{LEAD} for a triangulated or triaged site, \msg{CONVERGE} for a verified arrival, \msg{NOT FOUND} for a vetoed site, \msg{UNREACHED} for a site the robot could not reach, \msg{HELP}/\msg{GUIDE} for a route request and answer, and \msg{MISSION COMPLETE}. The body is free text from the LLM or a reflex template. The LLM reasons over every message; the reflex parses only tags and coordinates.

\section{Experimental Setup}
\label{sec:setup}

We evaluate HEROIC at two scales: a calibrated pure-Python abstraction for hundreds of paired episodes, and full-stack NVIDIA Isaac~Sim missions on procedurally generated scenes. 

\subsection{Simulator and Models}
Full-stack missions run in NVIDIA Isaac~Sim with PhysX dynamics and procedurally generated scenes with real colliders. Detector, VLM and LLM all run locally: GroundingDINO~\cite{liu2024groundingdino} in FP16 with one instance per robot, a CLIP false-positive classifier~\cite{radford2021clip}, and Qwen3-VL-8B served by vLLM as both the agent LLM and the VLM verifier. Missions are budgeted in \emph{simulation} time (900~s) so compute load cannot shorten a mission; the real-time factor is 0.3--0.4 on a 16~GB RTX~5080 that also hosts the simulator.

\subsection{Scene Tiers}
Seeded generators produce the tiers of Table~\ref{tab:tiers}; the agents see only the goal text and $R_s$. The standard and forest tiers are the searcher regime. The disaster tier motivates the work: a person inside a collapsed building, 30--55\% visible, with $\rmax$ of 4--8~m. The three tiers introduced with this paper each make a different mechanism the bottleneck: triage, ingress and multi-victim bookkeeping (warehouse); cue fusion under a canopy that blinds the air (hiker); building-level allocation from one cue (farm).


\begin{table}[t]
  \centering
  \caption{Scene tiers. $\rmax$ is for the goal class at the tier's visible fraction; the air agent is \emph{blind} when $\rmax$ is below the lowest safe altitude.}
  \label{tab:tiers}
  \scriptsize
  \setlength{\tabcolsep}{2.5pt}
  \begin{tabular}{@{}lp{5.0cm}l@{}}
    \toprule
    tier & target and what makes it hard & $\rmax$, regime \\
    \midrule
    standard & church, truck, person or bench on open ground with groves and depots & 7--122 m, searcher \\
    forest & hut under a random sparse/dense 24--30 m conifer canopy; detection-limited & 32 m, searcher \\
    disaster (USAR) & person inside a small 20$\times$20 m collapsed house, 30--55\% visible & 4--8 m, blind \\
    warehouse & 3--5 victims: 2--3 trapped in bays of a 40$\times$24 m collapsed warehouse (front door blocked, entries a 1.2 m crack and a collapsed corner), 1--2 in rubble outside & 4 m, blind \\
    hiker & person 4 m off a trail end, under dense canopy (380--480 stems/ha); breadcrumbs (backpack, notebook) on the ground & --, blind \\
    farm & person inside the farmhouse, the open barn or behind the shed; one cue (open door, parked truck, smoke) names the building & 5 m, blind \\
    \bottomrule
  \end{tabular}
\end{table}

\subsection{HEROIC-Lite: a Calibrated Abstraction}
HEROIC-Lite is a pure-Python abstraction of the same code paths: the same belief class, scale law, success criterion \eqref{eq:success} and baseline waypoint choosers, with the renderer and detector replaced by a detection model calibrated on 64k detections matched to ground truth in Isaac~Sim, so that $\eta(px)$, the false-positive rate and the range-dependent miss likelihood of \eqref{eq:neg} are those of the deployed stack. Lite is calibrated at the detector level only: it reproduces HEROIC's full-stack success within 8~pp on all three shared tiers, but it does not model structure ingress and route stalls, and baseline rates differ from Isaac~Sim by up to 56~pp. We therefore use Lite for ablations and paired significance, and Isaac~Sim for absolute baseline comparisons.

\subsection{Comparison Methods}
\label{sec:baselines}
Every baseline runs \emph{inside the same stack} with one function swapped: the idle-agent waypoint chooser. Baselines keep the detector, VLM verifier, ray belief, lead investigation, arrival gate, committed hold, \msg{CONVERGE}/\msg{NOT FOUND} protocol, air rendezvous and the scale-derived altitude and spacing of Sec.~\ref{sec:derived}; they do not get the LLM turn loop (the published methods have none), the role switch, aerial triage, the cue prior or salience ordering, and none reads ground truth. The co-arrival criterion required a bolt-on convergence rule for every baseline, disclosed in the code.
\begin{itemize}
    \item \textbf{Lawnmower:} a fixed boustrophedon at spacing $s$ over the disc (UAV in reverse order) that never reacts to observations; the coverage-complete reference.
    \item \textbf{Frontier}~\cite{yamauchi1997frontier}\textbf{:} frontier cells of the agent's own explored grid, thinned, with utility $=$ information gain $-$ travel cost $-$ teammate-claim penalty.
    \item \textbf{VLFM}~\cite{yokoyama2024vlfm}\textbf{:} the same frontiers chosen by a value map. The image--text similarity is replaced by goal-class detection mass near the frontier plus a class-affinity table over detected context objects. Disclosed as a proxy and deliberately generous, since the table encodes the scene generator's own placement priors.
    \item \textbf{SemGraph-LLM}~\cite{cladera2025airground,ravichandran2025spine}\textbf{:} the air--ground LLM planner of Cladera et al.\ with fixed roles. The UAV flies one coverage survey that populates the shared object graph and then shadows the UGV; the UGV asks the same local LLM to rank the graph's nodes against the mission text, visits them in order, then falls back to frontier exploration. No negotiation, no re-tasking, no co-arrival reasoning.
    \item \textbf{Random walk:} area-uniform goals in the disc; the floor.
    \item \textbf{HEROIC ablations} (same episodes): without the commit gate; without the ray belief (single-view depth points instead); without scale-aware altitude (the degenerate $a = \rmax$); without the air role switch (the UAV sweeps regardless of $\beta$).
    \item \textbf{HEROIC (proposed):} the full framework of Sec.~\ref{sec:method}.
\end{itemize}

\subsection{Evaluation Protocol}
\label{sec:protocol-eval}
Every comparison is \emph{paired on seeds}: each method runs on the same scene instances, with the same spawn, victim and cue placement, so that scene difficulty cancels. For success we use an exact McNemar test~\cite{mcnemar1947} on the discordant pairs, and for time-to-find a two-sided sign-flip permutation test with 20k resamples on the paired differences, failures censored at 900~s. Runs whose agents never acted (out-of-memory, frozen simulator) are re-run and never scored. In HEROIC-Lite this gives $n = 300$ paired episodes per condition and tier; in Isaac~Sim, 20--35 paired seeds per tier (158 in total), with success contrasts tested by exact McNemar on the pooled discordant pairs.

\section{Results}
\label{sec:results}

\subsection{Quantitative Comparisons to Baselines}

HEROIC has the highest success rate on all three tiers (Table~\ref{tab:lite}): 86.0\%, 76.3\% and 82.0\% on standard, forest and USAR against best-baseline rates of 74.0\%, 64.0\% and 65.7\%, a margin of $\sim$12~pp on the open tiers and up to 21~pp over frontier and VLFM; on USAR, frontier and random walk fall below 50\%. Every contrast is significant under a two-sided exact McNemar test~\cite{mcnemar1947} on the discordant pairs, computed under the pairing most unfavorable to HEROIC that is consistent with the marginals ($p<0.001$ throughout, except lawnmower on forest at $p<0.005$).

Time tells the same story more sharply (Table~\ref{tab:lite}). On the standard tier HEROIC's median mission of 129~s is 14--122~s shorter than every baseline's; on forest it is shorter than three of the four, and frontier's 233~s median comes with 16~pp fewer successes. On USAR HEROIC's median is 333~s; VLFM's is 743~s and frontier's and random walk's hit the censor because most of their episodes never find the victim. Where detectability is scarce HEROIC is not only more likely to reach the victim, it reaches them two to four times sooner, the metric that matters for triage.

\begin{table}[t]
  \centering
  \caption{HEROIC-Lite paired results, $n=300$ episodes per cell. Success in \%, median time-to-find in seconds. Bold: best per column. Lower block: ablations on same episodes, as change in success vs. full HEROIC discordant count (lost\,:\,gained).}
  \label{tab:lite}
  \scriptsize
  \setlength{\tabcolsep}{1.4pt}
  \begin{tabular}{@{}lrrr|rrr|rrr@{}}
    \toprule
     & \multicolumn{3}{c|}{standard} & \multicolumn{3}{c|}{forest} & \multicolumn{3}{c}{disaster (USAR)} \\
    method & succ. & med. & $\Delta t$ & succ. & med. & $\Delta t$ & succ. & med. & $\Delta t$ \\
    \midrule
    HEROIC    & \textbf{86.0} & \textbf{129} & -- & \textbf{76.3} & 280 & -- & \textbf{82.0} & \textbf{333} & -- \\
    VLFM      & 64.7 & 210 & $-81$ & 57.3 & 352 & $-72$ & 57.0 & 743 & $-410$ \\
    Frontier  & 67.3 & 189 & $-60$ & 59.7 & \textbf{233} & $+47$ & 44.0 & 900 & $-567$ \\
    Lawnmower & 72.0 & 251 & $-122$ & 64.0 & 308 & $-28$ & 65.7 & 516 & $-183$ \\
    Random    & 74.0 & 143 & $-14$ & 58.3 & 487 & $-207$ & 44.7 & 900 & $-567$ \\
    \midrule
    \multicolumn{10}{@{}l}{\emph{Ablations}: $\Delta$ success (pp), episodes lost\,:\,gained vs.\ HEROIC} \\
    $-$ commit gate      & $-74.0$ & \multicolumn{2}{c|}{226\,:\,4}  & $-63.0$ & \multicolumn{2}{c|}{191\,:\,2} & $-74.0$ & \multicolumn{2}{c}{223\,:\,1} \\
    $-$ ray belief       & $-32.7$ & \multicolumn{2}{c|}{122\,:\,24} & $-56.3$ & \multicolumn{2}{c|}{175\,:\,6} & $-12.0$ & \multicolumn{2}{c}{77\,:\,41} \\
    $-$ scale altitude   & $-23.7$ & \multicolumn{2}{c|}{85\,:\,14}  & $-0.7$  & \multicolumn{2}{c|}{2\,:\,0}   & $-1.0$  & \multicolumn{2}{c}{32\,:\,29} \\
    $-$ air role switch  & $0.0$   & \multicolumn{2}{c|}{0\,:\,0}    & $0.0$   & \multicolumn{2}{c|}{0\,:\,0}   & $-14.0$ & \multicolumn{2}{c}{42\,:\,0} \\
    \bottomrule
  \end{tabular}
\end{table}

\subsection{Component Ablations}

The lower block of Table~\ref{tab:lite} ablates the components on the same paired episodes. The \emph{commit gate} is worth 63--74~pp on every tier (226\,:\,4, 191\,:\,2 and 223\,:\,1 discordant): without verified commitment the team declares arrival at phantoms and never completes. The \emph{ray belief} is worth 33~pp on standard and 56~pp on forest, where far detections of large targets are common and their back-projected positions are wrong by tens of meters, and only 12~pp on USAR, where there are few far detections to mislead. \emph{Scale-aware altitude} is worth 23.7~pp on standard, where the air agent searches, and is inert (2\,:\,0 and 32\,:\,29) where it is blind. The \emph{air role switch} never fires on standard and forest (0\,:\,0 at $n=300$), where the UAV can already see the target, but is decisive on USAR: 42 episodes saved and none lost ($-14.0$~pp, $p<0.001$). The law thus finds the physical boundary at which the aerial role should change.

\subsection{Full-Stack Validation in Isaac Sim}
\label{sec:isaac}

\def\trajenv{figure}
\begin{\trajenv}[t]
  \centering
  \includegraphics[width=\linewidth]{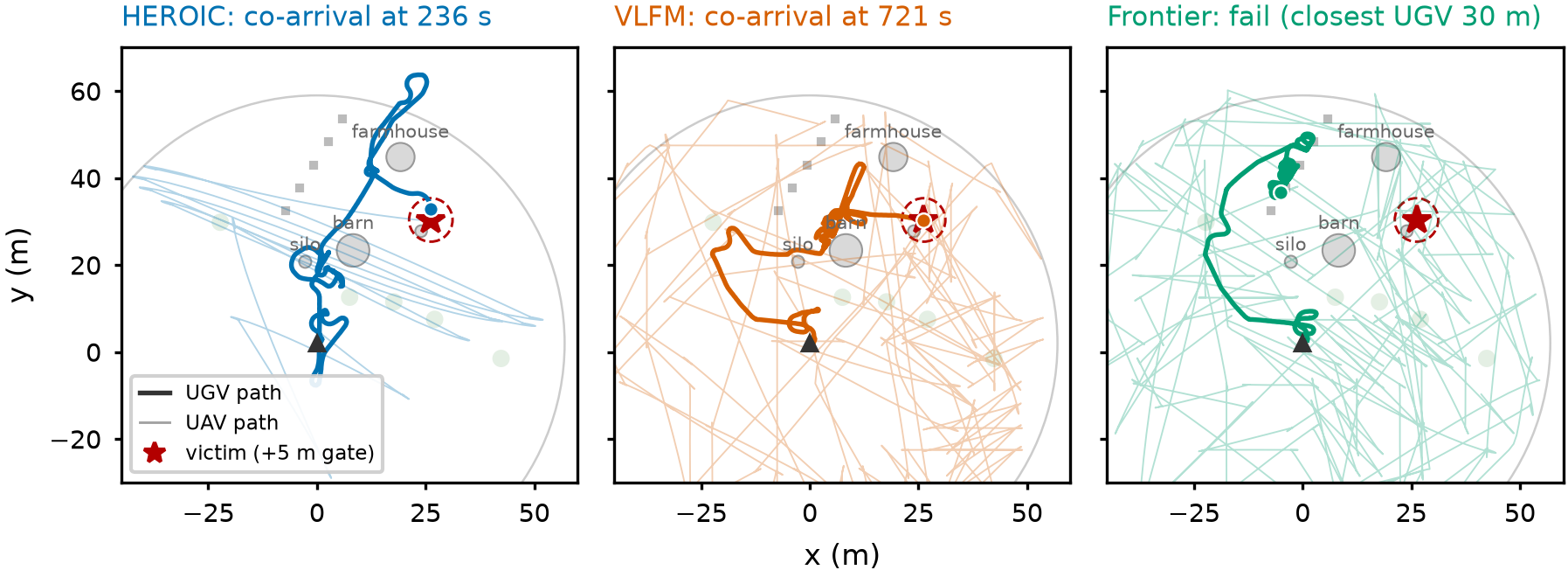}
  \caption{Isaac~Sim, one farm mission (victim behind the shed, smoke cue). Thick lines: UGV, thin lines: UAV, star: victim with its 5~m gate, triangle: spawn. HEROIC's UGV investigates the cued building and co-arrives at 236~s; VLFM's value map reaches the same place at 721~s; frontier exhausts the budget with the UGV 30~m away.}
  \label{fig:traj}
\end{\trajenv}

\begin{table}[t]
  \centering
  \caption{Isaac~Sim, full stack: success rate per tier and unweighted mean; bold is best per column.}
  \label{tab:isaac}
  \scriptsize
  \setlength{\tabcolsep}{2.0pt}
  \begin{tabular*}{\linewidth}{@{\extracolsep{\fill}}lccccccc@{}}
    \toprule
    method & hiker & farm & wareh. & disaster & forest & stand. & avg. \\
     & 25 & 35 & 34 & 20 & 24 & 20 & 6 tiers \\
    \midrule
    HEROIC       & 0.72 & \textbf{0.89} & \textbf{0.82} & \textbf{0.90} & 0.83 & \textbf{0.85} & \textbf{0.84} \\
    SemGraph-LLM & 0.16 & 0.51 & 0.03 & 0.40 & 0.67 & \textbf{0.85} & 0.44 \\
    VLFM         & 0.44 & 0.74 & 0.21 & 0.25 & \textbf{0.88} & 0.70 & 0.54 \\
    Frontier     & 0.68 & 0.31 & 0.00 & 0.15 & 0.75 & 0.65 & 0.42 \\
    Lawnmower    & \textbf{0.76} & 0.40 & 0.00 & 0.10 & 0.63 & 0.60 & 0.41 \\
    Random       & 0.44 & 0.00 & 0.03 & 0.20 & 0.75 & 0.70 & 0.35 \\
    \bottomrule
  \end{tabular*}
\end{table}

In the full-stack matrix (Table~\ref{tab:isaac}), the cue tiers reward context reasoning. On the farm tier HEROIC has the highest success rate at 0.89, with VLFM second at 0.74 and SemGraph-LLM third at 0.51; the methods that ignore context objects trail far behind, confirming that cues separate methods that use them from those that do not. HEROIC is also faster on the farm scene (236~s versus 721~s for VLFM, Fig.~\ref{fig:traj}). The warehouse tier is similar: HEROIC reaches 0.82; the next-best baseline, VLFM, reaches 0.21, while frontier and lawnmower fail entirely (0.00). The disaster tier separates the methods most sharply: HEROIC reaches 0.90 and no baseline exceeds SemGraph-LLM's 0.40.

The hiker tier exposes the limit of aerial assistance in dense canopies. The systematic lawnmower and frontier methods reach the hiker with success rates of 0.76 and 0.68 against HEROIC's 0.72, because the UAV cannot see the trail or the person through the canopy at any altitude. Under the sparser forest canopy, where the hut is resolvable from the air, HEROIC is comparable to VLFM.

HEROIC ties SemGraph-LLM on standard (both 0.85), consistent with the ablation finding that the role switch is inert on that tier. SemGraph-LLM's fixed roles fail on hiker and warehouse, where its UGV is a poor explorer, opening a 40~pp gap in the average. Averaged over all six tiers HEROIC leads every baseline by 30--49~pp; pooled over all 158 paired seeds every contrast is significant under an exact McNemar test ($p<0.001$).

\section{Discussion}
\label{sec:discussion}
HEROIC is, to our knowledge, the first air--ground team whose robots coordinate only in natural language for long horizon missions and whose division of labor is derived from sensor properties rather than declared. The scale law is a physical quantity computable from the mission prompt and camera intrinsics, so the role assignment needs neither scene knowledge nor a capability tag; the same framework that keeps the UAV searching for a church at $\rmax = 122$~m re-tasks it to triage, escort and guide for a buried victim at $\rmax = 4$--$8$~m. Keeping the slow language model out of the control loop is what makes a 4--8~B edge LLM a viable mission planner.

Several limitations qualify the claims. The headline statistics come from a calibrated abstraction of our own stack; Isaac~Sim confirms the sign of the contrasts on five of the six tiers. Everything is simulation: the detector, VLM and colliders are real, the renderer is not, and the locally served models only emulate on-robot LLM latency. The scale law needs a height for the goal noun and a visible fraction for occluded targets; today both come from the mission brief. Finally, the hiker tier shows that when the air agent has nothing to condition on, a systematic ground sweep is as good as ours; the cue chain reaches the map but not yet the sweep order. Future work will extend the one-bit role function to larger mixed teams, let the cue chain reorder the ground sweep, and move to real platforms with onboard compute and intermittent links.

\section{Conclusion}
\label{sec:conclusion}

HEROIC derives an air--ground team's division of labor from one measured detector property, verifies every arrival through an evidential belief and commit gate, and coordinates only in natural language, with no shared map. Across six simulated tiers it reaches the victim more often and 2--4$\times$ sooner than vision-language, semantic-graph and geometric baselines, and it is the only method that does not collapse when the air asset cannot see the target.


{
\bibliographystyle{IEEEtran}
\bibliography{references}
}

\end{document}